# Indowordnet's help in Indian Language Machine Translation

**Sreelekha S, Pushpak Bhattacharyya**
Indian Institute of Technology (IIT) Bombay, India
{sreelekha, pb}@cse.iitb.ac.in

**Abstract**

Being less resource languages, Indian-Indian and English-Indian language MT system developments faces the difficulty to translate various lexical phenomena. In this paper, we present our work on a comparative study of 440 phrase-based statistical trained models for 110 language pairs across 11 Indian languages. We have developed 110 baseline Statistical Machine Translation systems. Then we have augmented the training corpus with Indowordnet synset word entries of lexical database and further trained 110 models on top of the baseline system. We have done a detailed performance comparison using various evaluation metrics such as BLEU score, METEOR and TER. We observed significant improvement in evaluations of translation quality across all the 440 models after using the Indowordnet. These experiments give a detailed insight in two ways : (1) usage of lexical database with synset mapping for resource poor languages (2) efficient usage of Indowordnet sysnset mapping. More over, synset mapped lexical entries helped the SMT system to handle the ambiguity to a great extent during the translation.

## 1. Introduction

Machine Translation (MT) faces difficulty when dealing with morphologically complex languages. Being a country with rich linguistic diversity, India has 22 scheduled languages and 30 Indian languages. These languages are spread across four language families such as; Indo-Aryan, Dravidian, Tibeto-Burman and Austro-Asiatic with 10 major scripts. Out of these, Hindi is the most prominent, which belongs to the Indo-Aryan family of languages. Most of the official documents are either in Hindi or English. 95% of the population is illeterate of English. Thus, for a proper functioning, there is a large requirement to translate these official documents into regional languages. More over, the medias and news agencies are required to translate news received in English from International news agencies to respective regional languages. Hence, there is a huge requirement for automatic MT system developments between English to Indian languages and Indian to Indian languages. To handle this lingusitistic diversity and rich morphology with lack of proper resources is the major challenge faced during the development of MT system between Indian languages. There were many MT system developments are going on for Indian languages using rule-based, statistical-based and hybrid approaches (Antony P. J. 2013; Ashan et. al., 2010; Brown et. al., 1993; Nair, et.al., 2012; Sreelekha et. al., 2013; Sreelekha et. al., 2015; Sreelekha et. al., 2017; Sreelekha et. al., 2018). Out of these, Statical MT(SMT) approach is the most promising due to its flexibility and it's easiness to develop. In this work, we developed phrase-based SMT systems for 110 language pairs and our further attempts to improve the quality of the translation systems on top of these baseline systems. After analyzing the developed SMT systems, we observed that the system fails to handle various linguistic phenomena and inflected word forms. Hence, we have decided to use the Indowordnet for SMT system development as a lexical database, which covers, dictionary words, transliteration, short phrases and coined words.

## 2. Indowordnet

IndoWordnet (Bhattacharyya, 2010) is as lexical database for various Indian languages, in which Hindi wordnet is the root and all other Indian language wordnets are linked through the expansion approach. Words and it's concepts are stored in a structure called the Lexical Matrix, where rows represent *word meanings* and columns represents the *forms*. IndoWordnet stores different words and relations mainly Lexical Relations and Semantic Relations. Different types of Lexical Relations such as Gradation for state, size, light, gender, temperature, color, time, quality, action, manner, Antonymy for action, amount, direction, gender, personality, place, quality, size, state, time, color, manner, Compound for nouns and Conjunction for verbs. Semantic Relation types such as Hypernymy for noun and verbs, Holonymy for nouns, Meronymy for component object, member collection, feature, activity, place, area, face, state, portion, mass, resource, process, position, area, Troponymy for verbs, Similar Attribute between noun and adjective, Function verb between noun and verb, Ability verb between noun and verb, Capability verb between noun and verb, Adverb modifies verb between adverb and verb, Causative for verb, Entailment for verb, Near synset and Adjective modifies noun between adjective and noun.

There are three principles the synset construction process must adhere to. Minimality principle insists on capturing that minimal set of the words in the synset which uniquely identifies the concept. For example *{family, house}* uniquely identifies a concept (*e.g. "he is from the house of the King of Jaipur"}*. Coverage principle then stresses on the completion of the synset, *i.e.,* capturing ALL the words that stand for the concept expressed by the synset (*e.g., {family, house, household, ménage} completes the synset*). Within the synset, the words will be ordered according to their frequency in the corpus. Replaceability demands that the most common words in the synset, *i.e.*, words towards the beginning of the synset should be able to replace one another in the example sentence associated with the synset. The comparative performance analysis with phrase based SMT models with that of augmented lexical database is described in Section 3 & 4.

## 3. Experimental Discussion

We discuss the experiments conducted on our English-Indian language and Indian-Indian language SMT systems for 11 Indian languages. We have conducted various experiments on a combination of 110 language pair using the Indian Language Corpora Initiative (ILCI) corpus (Choudhary and Jha, 2011 ; Jha, 2012) with 50000 parallel sentences. We trained 440 different models on top of the baseline SMT system. We have augmented the extracted Indowordnet synset words into the training corpus for the 110 language pairs. We describe the resources and the comparisons of results in the form of an error analysis. We have used Moses (Koehn et. al., 2007) and Giza++[1] for modeling the baseline system.

An algorithm is used to extract the bilingual mapped words from IndoWordnet. These entries have been entered manually into the system by linguistic experts with qualification of Masters in Literature in the respective languages over a period of 10 years. Bilingual mappings are generated using the concept-based approach across words and synsets (Mohanty et.al., 2008). The extracted entries have been validated manually. For a single word considered it’s all synset word mappings and generated that many entries of parallel words. For example, the word *अंतहीन {antaheen}{endless}* has the following equivalent synset words in IndoWordnet.

*अंतहीन: अनंत असमाप्य अंतहीन अनन्त अन्तहीन अनवसान*
*{antaheen: anantu asamapya antaheen anant antaheen anavasaan} {endless: endless not-ending endless infinite endless not-ending }*
The experiments conducted are as follows: Baseline SMT system with a cleaned corpus; SMT system with IndoWordnet extracted words. Consider an example from the English-Malayalam SMT system.

For example, English Sentence,

*He blow up the things.*

The Baseline English-Malayalam SMT system translated the above sentence in Malayalam as, *അവൻ അടിക്കുക കാര്യങ്ങൾ{avan adikkuka karyam} {He beat things}*

Here the system fails to translate the meaning of “*blow up*” correctly as a single word.

Then we have added the synset word entries of *blow_up* from the Indowordnet to the training corpus. The word ***blow_up*** has the following equivalent synset words in the IndoWordnet. ***blow_up:*** *വികസിപ്പിക്കുക വലുതാക്കുക വീറ്പ്പിക്കുക. {blow_up: vikasippikkuka valuthakkuka veerppikkuka} {blow_up: make-enlarge make-big make-full}*

| | Ben | Eng | Guj | Hin | Kon | Mal | Mar | Pan | Tam | Tel | Urd | |
|---|---|---|---|---|---|---|---|---|---|---|---|---|
| Ben | | 18.34<br>21.76 | 31.24<br>33.21 | 36.16<br>38.95 | 23.16<br>26.68 | 8.62<br>10.01 | 19.79<br>22.87 | 31.84<br>33.12 | 8.88<br>9.45 | 13.18<br>15.21 | 24.91<br>26.32 | WOW<br>WWN |
| Eng | 14.85<br>16.75 | | 17.36<br>19.83 | 26.53<br>29.45 | 13.01<br>15.67 | 4.85<br>6.84 | 10.17<br>13.54 | 22.86<br>26.78 | 4.17<br>6.32 | 6.43<br>8.67 | 18.07<br>21.09 | WOW<br>WWN |
| Guj | 29.35<br>32.89 | 17.45<br>22.65 | | 53.09<br>58.98 | 29.63<br>35.45 | 7.97<br>8.89 | 26.99<br>31.78 | 47.60<br>51.89 | 9.97<br>11.37 | 16.57<br>18.08 | 34.77<br>38.45 | WOW<br>WWN |
| Hin | 36.31<br>39.32 | 28.15<br>31.23 | 53.29<br>55.26 | | 36.06<br>39.79 | 10.95<br>12.87 | 33.78<br>39.87 | 70.06<br>75.68 | 11.36<br>19.67 | 21.59<br>27.65 | 50.30<br>56.43 | WOW<br>WWN |
| Kon | 24.61<br>27.45 | 16.92<br>18.93 | 31.44<br>35.56 | 36.54<br>41.67 | | 8.05<br>10.65 | 23.69<br>29.32 | 33.53<br>37.42 | 8.96<br>11.26 | 13.40<br>16.56 | 23.54<br>26.56 | WOW<br>WWN |
| Mal | 7.01<br>10.01 | 8.36<br>15.923 | 10.99<br>15.89 | 14.50<br>19.67 | 9.36<br>15.56 | | 7.01<br>10.01 | 08.53<br>12.65 | 6.67<br>10.89 | 6.25<br>10.01 | 08.27<br>12.76 | WOW<br>WWN |
| Mar | 22.68<br>26.68 | 14.87<br>18.45 | 33.84<br>39.54 | 41.66<br>52.67 | 27.44<br>36.98 | 7.25<br>10.54 | | 34.75<br>38.56 | 7.34<br>10.08 | 12.02<br>16.57 | 25.08<br>29.98 | WOW<br>WWN |
| Pan | 30.27<br>36.98 | 24.01<br>29.56 | 46.24<br>49.56 | 71.26<br>77.89 | 26.44<br>29.98 | 7.49<br>10.65 | 25.54<br>29.67 | | 8.96<br>12.67 | 17.92<br>20.67 | 44.46<br>53.67 | WOW<br>WWN |
| Tam | 12.77<br>15.78 | 10.90<br>15.98 | 17.28<br>21.90 | 21.79<br>25.89 | 14.17<br>19.34 | 6.41<br>12.48 | 11.10<br>16.45 | 19.32<br>25.78 | | 13.92<br>19.45 | 15.65<br>19.54 | WOW<br>WWN |
| Tel | 16.87<br>20.87 | 12.09<br>17.89 | 22.22<br>26.87 | 27.20<br>35.87 | 16.98<br>20.87 | 6.58<br>11.34 | 13.47<br>18.45 | 25.14<br>29.89 | 7.49<br>13.67 | | 19.03<br>25.87 | WOW<br>WWN |
| Urd | 26.14<br>31.67 | 21.00<br>28.56 | 38.92<br>45.78 | 58.09<br>67.54 | 25.09<br>30.87 | 7.49<br>13.24 | 21.21<br>29.56 | 51.90<br>59.67 | 8.13<br>13.78 | 14.65<br>20.67 | | WOW<br>WWN |

**Table 1: BLEU evaluation scores with and without Indowordnet**

[1] http://www.statmt.org/

| | Ben | Eng | Guj | Hin | Kon | Mal | Mar | Pan | Tam | Tel | Urd | |
|---|---|---|---|---|---|---|---|---|---|---|---|---|
| Ben | | 69.319<br>63.051 | 48.267<br>44.791 | 45.313<br>41.341 | 59.243<br>55.654 | 83.703<br>80.231 | 65.61<br>62.871 | 49.901<br>44.321 | 78.411<br>77.112 | 75.851<br>73.543 | 60.002<br>53.789 | WOW<br>WWN |
| Eng | 73.623<br>71.678 | | 68.464<br>64.983 | 57.623<br>54.895 | 76.350<br>72.541 | 93.146<br>91.043 | 83.371<br>80.345 | 62.606<br>59.789 | 92.524<br>90.897 | 90.670<br>88.981 | 70.185<br>65.876 | WOW<br>WWN |
| Guj | 51.77<br>45.678 | 69.561<br>63.219 | | 29.121<br>26.984 | 51.619<br>48.985 | 84.253<br>82.456 | 56.503<br>50.361 | 34.666<br>30.435 | 81.378<br>78.345 | 69.058<br>65.345 | 48.987<br>42.567 | WOW<br>WWN |
| Hin | 45.238<br>41.233 | 54.938<br>52.456 | 29.017<br>28.012 | | 46.649<br>41.982 | 84.488<br>88.456 | 51.305<br>39.87 | 18.079<br>16.89 | 78.704<br>71.78 | 66.396<br>62.567 | 37.525<br>33.561 | WOW<br>WWN |
| Kon | 57.462<br>55.871 | 69.003<br>66.931 | 48.567<br>47.834 | 43.953<br>40.532 | | 88.070<br>86.765 | 60.401<br>58.981 | 47.944<br>42.871 | 88.740<br>76.764 | 74.498<br>70.432 | 57.056<br>53.781 | WOW<br>WWN |
| Mal | 88.209<br>85.432 | 78.141<br>73.789 | 73.422<br>69.012 | 68.343<br>62.435 | 75.119<br>66.912 | | 85.472<br>81.875 | 79.721<br>74.872 | 96.695<br>91.034 | 89.002<br>81.098 | 75.498<br>71.016 | WOW<br>WWN |
| Mar | 57.868<br>51.789 | 68.783<br>62.832 | 45.651<br>41.765 | 40.611<br>31.984 | 54.779<br>49.789 | 85.370<br>81.870 | | 45.392<br>41.678 | 79.183<br>73.567 | 75.723<br>68.056 | 57.717<br>53.670 | WOW<br>WWN |
| Pan | 51.961<br>45.870 | 60.727<br>53.987 | 35.210<br>30.679 | 17.424<br>15.456 | 52.693<br>48.678 | 89.171<br>81.785 | 59.596<br>51.765 | | 80.315<br>75.897 | 69.862<br>64.467 | 41.548<br>32.653 | WOW<br>WWN |
| Tam | 71.499<br>67.789 | 78.701<br>70.657 | 64.033<br>60.432 | 63.700<br>56.890 | 70.220<br>65.789 | 88.092<br>82.945 | 77.660<br>71.670 | 64.957<br>60.345 | | 69.862<br>59.076 | 69.167<br>61.650 | WOW<br>WWN |
| Tel | 66.351<br>61.945 | 72.422<br>63.567 | 58.236<br>50.678 | 52.959<br>43.678 | 65.298<br>59.543 | 86.061<br>80.678 | 72.877<br>66.834 | 55.937<br>46.934 | 79.616<br>70.345 | | 64.792<br>55.897 | WOW<br>WWN |
| Urd | 58.339<br>48.234 | 64.693<br>58.456 | 44.488<br>38.567 | 29.017<br>21.765 | 60.469<br>51.789 | 90.603<br>82.567 | 67.468<br>59.378 | 34.060<br>27.456 | 84.165<br>76.457 | 77.524<br>68.456 | | WOW<br>WWN |

**Table 2: TER Evaluation scores for with and without Indowordnet**

| | Ben | Eng | Guj | Hin | Kon | Mal | Mar | Pan | Tam | Tel | Urd | |
|---|---|---|---|---|---|---|---|---|---|---|---|---|
| Ben | | 0.243<br>0.268 | 0.526<br>0.531 | 0.570<br>0.591 | 0.433<br>0.439 | 0.229<br>0.241 | 0.396<br>0.405 | 0.510<br>0.520 | 0.262<br>0.271 | 0.292<br>0.308 | 0.447<br>0.456 | WOW<br>WWN |
| Eng | 0.320<br>0.330 | | 0.375<br>0.383 | 0.496<br>0.513 | 0.306<br>0.319 | 0.153<br>0.165 | 0.279<br>0.284 | 0.435<br>0.458 | 0.191<br>0.215 | 0.201<br>0.212 | 0.391<br>0.402 | WOW<br>WWN |
| Guj | 0.483<br>0.499 | 0.372<br>0.387 | | 0.714<br>0.737 | 0.502<br>0.527 | 0.218<br>0.237 | 0.480<br>0.492 | 0.650<br>0.672 | 0.287<br>0.296 | 0.348<br>0.351 | 0.541<br>0.555 | WOW<br>WWN |
| Hin | 0.559<br>0.564 | 0.315<br>0.321 | 0.732<br>0.745 | | 0.583<br>0.596 | 0.277<br>0.285 | 0.574<br>0.775 | 0.826<br>0.878 | 0.214<br>0.318 | 0.339<br>0.398 | 0.576<br>0.673 | WOW<br>WWN |
| Kon | 0.428<br>0.449 | 0.229<br>0.242 | 0.524<br>0.541 | 0.592<br>0.610 | | 0.119<br>0.123 | 0.381<br>0.398 | 0.522<br>0.545 | 0.260<br>0.271 | 0.298<br>0.318 | 0.450<br>0.481 | WOW<br>WWN |
| Mal | 0.234<br>0.246 | 0.147<br>0.156 | 0.273<br>0.287 | 0.316<br>0.335 | 0.234<br>0.249 | | 0.194<br>0.209 | 0.255<br>0.268 | 0.179<br>0.189 | 0.176<br>0.187 | 0.193<br>0.205 | WOW<br>WWN |
| Mar | 0.420<br>0.446 | 0.219<br>0.245 | 0.543<br>0.587 | 0.617<br>0.854 | 0.489<br>0.518 | 0.212<br>0.230 | | 0.534<br>0.581 | 0.253<br>0.269 | 0.282<br>0.306 | 0.446<br>0.458 | WOW<br>WWN |
| Pan | 0.491<br>0.593 | 0.284<br>0.312 | 0.665<br>0.721 | 0.831<br>0.858 | 0.513<br>0.530 | 0.229<br>0.238 | 0.480<br>0.598 | | 0.232<br>0.243 | 0.370<br>0.389 | 0.626<br>0.745 | WOW<br>WWN |
| Tam | 0.244<br>0.287 | 0.180<br>0.192 | 0.365<br>0.386 | 0.402<br>0.421 | 0.313<br>0.332 | 0.189<br>0.198 | 0.267<br>0.284 | 0.362<br>0.387 | | 0.270<br>0.312 | 0.329<br>0.343 | WOW<br>WWN |
| Tel | 0.336<br>0.357 | 0.189<br>0.210 | 0.416<br>0.434 | 0.470<br>0.498 | 0.350<br>0.376 | 0.192<br>0.219 | 0.303<br>0.320 | 0.434<br>0.465 | 0.236<br>0.251 | | 0.369<br>0.389 | WOW<br>WWN |
| Urd | 0.455<br>0.475 | 0.271<br>0.321 | 0.602<br>0.623 | 0.752<br>0.812 | 0.463<br>0.486 | 0.218<br>0.234 | 0.431<br>0.452 | 0.696<br>0.756 | 0.260<br>0.283 | 0.327<br>0.365 | | WOW<br>WWN |

**Table 3: METEOR Evaluation scores for with and without Indowordnet**

After augmenting the synsets of ***blow up*** to the corpus, the SMT system with IndoWordnet model was able to translate the above English sentence correctly in *Malayalam as, അവൻ കാര്യങ്ങൾ വലുതാക്കുന്നു {avan karyangal valuthakkunnu} {He blow up things }* Since the synsets covers all common forms of a word, the augmentation of extracted parallel synset words in to the training corpus not only helped in improving the translation quality to a great extent but also helped in handling the word sense disambiguation well. We have added indowordnet synset entries for the entire 110 language pairs like this way and conducted the comparative experiments.

## 4. Evaluation & Error Analysis

We have used a tuning (MERT) corpus of 500 sentences from ILCI corpus. We have tested the translation system with 1000 sentences taken from the ILCI corpus. We have evaluated the translated outputs of all the 440 SMT systems using various evaluation methods such as BLEU score (Papineni et al., 2002), METEOR and TER (Agarwal and Lavie 2008) to analyze better. The results are shown in Tables 1, 2 and 3. There are two entries on each cell. The first row is showing the baseline SMT system results and it is represented as With-Out-Wordnet (WOW) at the last column of each row. The second row of each cell is representing the scores of baseline SMT system with Indowordnet's lexical entries experiments and it is represented as With-Wordnet (WWN) at the last column of each row. Table 1 shows the BLEU evaluation score for both the Baseline SMT system and the SMT system with Indowordnet lexical entries. Table 3 shows the METEOR evaluation score for both the Baseline SMT system and the SMT system with Indowordnet lexical entries. Table 2 shows TER evaluation score for both the Baseline SMT system and the SMT system with the Indowordnet lexical entries. We couldn't include the tables of without tuning experimental results due to the page limit. We observed that, the quality of the translation is improving with the usage of Indowordnet lexical databse. Hence, there is an incremental growth in BLEU score, METEOR score and reduction in TER score.

## 5. Conclusion

In this paper we have mainly focused on the usage of Indowordnet lexical entries for improving the quality of Indian-Indian and English-Indian language Machine Translation. We have discussed the comparative performance of phrase based baseline SMT system for the 110 language pairs and further the SMT system with Indowordnet entries on top of the baseline model. As discussed in the experimental Section, SMT system's translation quality has improved significantly with the usage of Indowordnet lexical entries. Moreover, the system was able to handle the rich morphological inflections and ambiguity to a great extend. We can see that there is an incremental growth in terms of BLEU-Score, METEOR and a decrement of TER evaluations, which shows the translation quality improvement. This leads to the importance of utilizing wordnet lexical resources for developing an efficient Machine Translation system for morphologically rich languages.

## Acknowledgements

This work is funded by Dept. of Science and Technology, Govt. of India under Women Scientist Scheme- WOS-A with the project code- SR/WOS-A/ET-1075/2014, TDIL.